\documentclass[unnumsec,webpdf,contemporary,large]{oup-authoring-template}%

\graphicspath{{Fig/}}

\theoremstyle{thmstyleone}%
\theoremstyle{thmstyletwo}%
\theoremstyle{thmstylethree}%

\begin{document}

\journaltitle{Journal Title Here}
\DOI{DOI HERE}
\copyrightyear{2022}
\pubyear{2019}
\access{Advance Access Publication Date: Day Month Year}
\appnotes{Paper}

\firstpage{1}


\title[MLMF]{Multi-layer matrix factorization for cancer subtyping using full and partial multi-omics dataset}

\author[1]{Yingxuan Ren}
\author[2]{Fengtao Ren}
\author[3,4,$\ast$]{Bo Yang}

\authormark{Ren et al.}

\address[1]{\orgdiv{Department of Computer Science}, \orgname{The University of Hong Kong}, \state{Hong Kong}, \country{China}}
\address[2]{\orgdiv{Department of engineering}, \orgname{The Chinese University of Hong Kong}, \state{Hong Kong}, \country{China}}
\address[3]{\orgdiv{School of Computer Science}, \orgname{Xi’an Polytechnic University}, \postcode{710048}, \state{ Xi’an}, \country{China}}
\address[4]{\orgdiv{Donnelly Centre for Cellular and Biomolecular Research}, \orgname{University of Toronto}, \postcode{ON M5S 3E1}, \state{Toronto}, \country{Canada}}

\corresp[$\ast$]{Corresponding author. \href{email:email-id.com}{yangboo@stu.xjtu.edu.cn}}

\received{Date}{0}{Year}
\revised{Date}{0}{Year}
\accepted{Date}{0}{Year}



\abstract{Cancer, with its inherent heterogeneity, is commonly categorized into distinct subtypes based on unique traits, cellular origins, and molecular markers specific to each type. However, current studies primarily rely on complete multi-omics datasets for predicting cancer subtypes, often overlooking predictive performance in cases where some omics data may be missing and neglecting implicit relationships across multiple layers of omics data integration. This paper introduces Multi-Layer Matrix Factorization (MLMF), a novel approach for cancer subtyping that employs multi-omics data clustering. MLMF initially processes multi-omics feature matrices by performing multi-layer linear or nonlinear factorization, decomposing the original data into latent feature representations unique to each omics type. These latent representations are subsequently fused into a consensus form, on which spectral clustering is performed to determine subtypes. Additionally, MLMF incorporates a class indicator matrix to handle missing omics data, creating a unified framework that can manage both complete and incomplete multi-omics data. Extensive experiments conducted on 10 multi-omics cancer datasets, both complete and with missing values, demonstrate that MLMF achieves results that are comparable to or surpass the performance of several state-of-the-art approaches.}


\maketitle

\section{Introduction}

Cancer is one of the major global health threats, with its high incidence and mortality rates making it a focal point of current medical research and public health efforts. Its occurrence and development are a biological change with a complex mechanism. Different subtypes of the same cancer may differ in histopathology and clinical features, but the heterogeneity of cancer mainly stems from its intrinsic molecular characteristics \citep{reis2011gene}. Therefore, making full use of the intrinsic molecular characteristics of cancer to identify cancer subtypes will help achieve precision medicine for cancer. In precision medicine, the molecular profile of a patient contains multiple molecules that belong to different omics (such as genomics, proteomics, metabolomics, etc.). These omics data reflects different biological processes, such as gene expression, protein function, metabolic pathways, etc. Early studies usually conducted statistics and research on one single omics data \citep{sotiriou2003breast}. However, one single omics data can only reflect the cancer characteristics of a certain level of biological process \citep{etcheverry2010dna}, and using different single omics data to address the same question may produce different results. For example, using mRNA expression data and Copy Number Variation (CNV) data to identify the subtype of breast cancer samples, the identification results are significantly different \citep{burgun2008accessing}. Incompatible subtype classifications cannot have a positive effect on clinical treatment. For a heterogeneous disease like cancer, its occurrence and development are affected by different gene combinations and various factors, so only using single omics data cannot fully describe the complete information of cancer \citep{cai2024deeply}. Different omics data are combined to describe the patient's biological information, which is called "multi-omics data" \citep{subramanian2020multi}. Currently, common multi-omics data includes CNV, mRNA expression, miRNA expression, DNA methylation, etc. \citep{shahrajabian2023survey}. Multi-omics data can reflects the various biological processes in cancer. Effective mining and integration of multi-omics data can effectively make up for the shortcomings of single-omics data, thereby comprehensively understanding the occurrence and development of cancer \citep{kumar2024next}. 

Currently, cancer subtype identification based on multi-omics data is mainly achieved through the integrated analysis of cancer sample data \citep{yang2022mdicc}. With the widespread application of machine learning, such as multi-view learning and deep learning, the current methods can be roughly divided into three categories: early integration, mid-term integration, and late integration \citep{ma2022mocsc}. For early integration, the main principle is to concatenate the input feature matrices of different omics into a multi-omics feature matrix, and then apply traditional clustering algorithms such as K-means, spectral clustering, etc. on the multi-omics feature matrix \citep{chen2023deep}. Through clustering, each category corresponds to a different cancer subtype. For example, A Bayesian latent model \citep{lock2013bayesian} simultaneously finds the latent low-dimensional subspaces and assigns samples into different clusters, so that different clusters represent different cancer subtypes. LRAcluster \citep{wu2015fast} is an integrated probability model based on low-rank approximation. It finds the global optimal solution of the objective function through a simple gradient ascent algorithm, and then uses the K-means method on the latent representation matrix to obtain the results of cancer subtypes \citep{duan2021evaluation}. For early integration,  data fusion is achieved by direct splicing, hence the integrating process cannot reflect on the correlation between different omics. However, due to overly simple operations, the spliced data often contains redundant information, which increases the data dimension of the input model. The main principle of late integration is to use the clustering algorithm of a single omics on each omics separately, and then integrate the different clustering results obtained from all omics as the final identification  result \citep{yuanyuan2021ssig}. The PINS method \citep{nguyen2017novel} constructs a connectivity matrix by integrating the clustering results of various omics data and integrates the connectivity matrix into a similarity matrix for clustering. The CC algorithm \citep{monti2003consensus} verifies the rationality of clustering by randomly extracting subsets from the original data, specifying the number of clusters, and clustering all data subsets separately. Although the late integration method does not increase the data dimension of the input model, it can adopt a single omics normalization for each data type and use a model adapted to each omics data, but it cannot establish inter-omics associations at the feature level. Mid-term integration is the most common mainstream method. The MCCA algorithm \citep{witten2009extensions} uses sparse canonical correlation analysis to find highly correlated omics data. iClusterBayes \citep{mo2013pattern} based on iCluster uses a full Bayesian latent variable model to select valuable latent variables and describe the intrinsic structure in multi-omics data. Xu et al. \citep{xu2021network} proposed the MSNE algorithm to integrate multi-omics information by embedding similarity relationships of samples defined by random walks on multiple similarity networks. 

Another problem with using multi-omics data to identify cancer subtypes is that the high cost of sequencing technology can lead to incomplete multi-omics data. Some patients may only have their mRNA expression data or DNA methylation data sequenced. In this case, there is no complete available multi-omics data. If a complete clustering algorithm based on multi-omics data is used in incomplete samples, it will inevitably fail and affect the performance of clustering. For incomplete data, a common method is to delete all samples with missing data and only consider samples with complete data. Obviously, this strategy will reduce the number of samples and waste hard-earned data. Another strategy is to use the KNN interpolation method \citep{troyanskaya2001missing} to fill in missing data, but the filled data may have a negative impact on the clustering results \citep{shi2022multiview}. Some methods proposed recently have begun to address the problem of incomplete data. NEMO \citep{rappoport2019nemo} allows samples to be missing in one or more datasets. If each pair of samples has a measurement value in at least one omics data set, cancer subtypes can be identified. MSNE \citep{xu2021network} captures the comprehensive similarity of samples by random walks on multiple similarity networks and is also applicable when data is missing. Therefore, how to effectively use these incomplete multi-omics data to better identify cancer subtypes has become an important issue in this research field.

Therefore, a Multi-Layer Matrix Factorization method called MLMF, for cancer subtyping via multi-omics data clustering is proposed in this paper. MLMF first takes the feature matrix of multi-omics as input, performs multi-layer linear or nonlinear factorization on the matrix, decomposes the original multi-omics data representation into their respective latent feature representations, and then fuses these representations into a consensus representation. Finally, spectral clustering is performed on this consensus representation. In addition, an indicator matrix is used to represent the missing status of some samples in the omics, thereby unifying the processes of complete multi-omics and missing multi-omics in a common framework.

\section{Method}\label{sec2}

\begin{figure*}[!t]%
\centering
{\includegraphics[width=\textwidth]{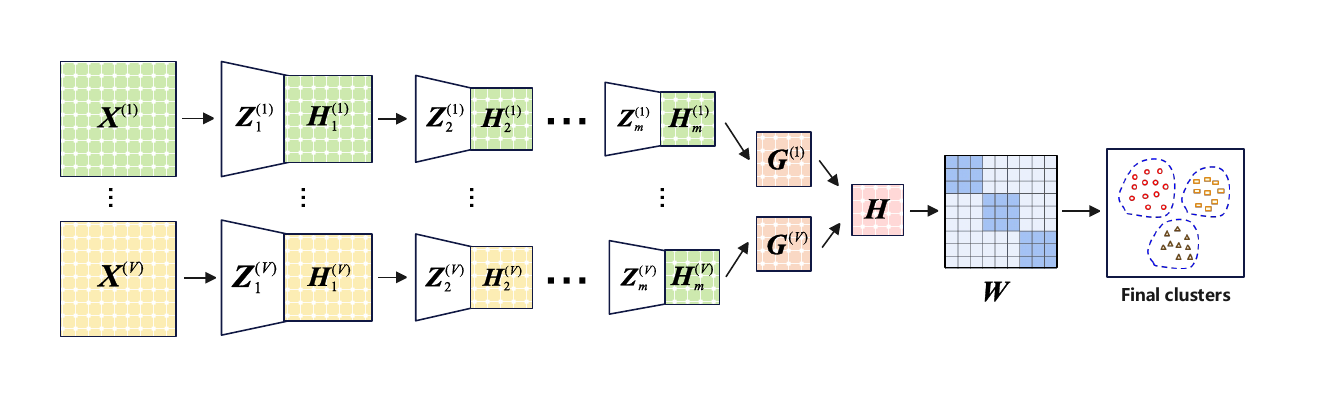}}
\caption{The framework of MLMF. MLMF is an m-layer matrix decomposition structure based on multi-omics data. It is an iterative process that can decompose each omics data matrix $\boldsymbol{X}^{(v)}$ into two factor matrices ($\boldsymbol{Z}_i^{(v)}$,$\boldsymbol{H}_i^{(v)}$), and then
fuses these factor matrices into a consensus representation $\boldsymbol{H}$, and optimized use two different cost functions, linear decomposition and nonlinear decomposition. Finally, cancer subtype is identified on consensus representation $\boldsymbol{H}$ via spectral clustering.}
\label{fig1}
\end{figure*}

MLMF mainly includes two modules, i.e. matrix factorization and optimizing consensus. Cancer subtyping is carried out on the consensus representation via spectral clustering algorithm. Each module and step will be detailed in the following sections.

\subsection{Notation}\label{subsec1}
Let $\boldsymbol{X}=\{\boldsymbol{X}^{(1)},\boldsymbol{X}^{(2)},\ldots,\boldsymbol{X}^{(V)}\}$ represents multi-omics dataset where $V$ is the number of omics. $\boldsymbol{X}^{(v)}=\left\{\boldsymbol{x}_1^{(v)},\boldsymbol{x}_2^{(v)},...,\boldsymbol{x}_{N_v}^{(v)}\right\}\in\mathbb{R}^{D_v\times N_v}$ is a collection of $N_v$ data samples with dimension $D_v$ in $vth$ omics measurements, where $v=1,2,\ldots,V$. The consensus representation is $\boldsymbol{H}=\left\{\boldsymbol{h}_1,\boldsymbol{h}_2,\ldots\boldsymbol{h}_N\right\}^T\in\mathbb{R}^{N\times d}$, where $d$ is the ultimate dimension of consensus embedding space and $N\ (N\geq N_v)$ is the sample size of total data. ${\parallel\cdot\parallel}_F^2$ is the Frobenius norm.

Since data may be missing, the sample index matrix $\boldsymbol{G}^{\left(v\right)}$ on each omics data is constructed as follows:
\begin{equation}
\small
\boldsymbol{G}_{ij}^{(v)} = \begin{cases}
1, & \text{if } i\text{th sample in } \boldsymbol{X}^{(v)} \text{ is the } j\text{th sample in intact data} \\
0, & \text{otherwise}
\end{cases}
\label{eq1}
\end{equation}

\subsection{The framework of MLMF}\label{subsec1}

As shown in  Figure \ref{fig1}, MLMF mainly contains two modules. First, the deep semi-non-negative matrix factorization algorithm is used to perform multi-layer factorization of each omics data to obtain a deep low-dimensional representation. According to the mapping way, it can be formulated two strategies: linear mapping and nonlinear mapping. Then in
the consensus representation module, indicator matrix is used to
represent the missing status of some samples in the omics, and then fuses these representations into a consensus representation. The consensus representation retains as much original information as possible through the minimum reconstruction loss. Finally, cancer subtype is identified on consensus representation via spectral clustering.

\subsection{Deep semi-non-negative matrix factorization}\label{subsec2}

Non-negative matrix factorization (NMF) \citep{han2015non} is a classic matrix factorization algorithm that adds non-negative constraints to matrix factorization and maps non-negative original data into a low-dimensional space consisting of many non-negative vectors. To extend the applicability of NMF, semi-nonnegative matrix factorization (Semi-NMF) \citep{ding2008convex} was introduced, which is a variant of NMF that allows the data matrix to be not strictly non-negative. Suppose the original data $\boldsymbol{P}=[\boldsymbol{p}_1,\boldsymbol{p}_2,...,\boldsymbol{p}_n]\in\mathbb{R}^{D\times N}$, so it could be decomposed into the basis matrix $\boldsymbol{O}=\begin{bmatrix}\boldsymbol{o}_1,\boldsymbol{o}_2,...,\boldsymbol{o}_n\end{bmatrix}\in\mathbb{R}^{D\times k}$ and the coefficient matrix $\boldsymbol{Q}=[\boldsymbol{q}_1,\boldsymbol{q}_2,...,\boldsymbol{q}_n]\in\mathbb{R}^{k\times N}$. Semi-NMF only imposes non-negative constraints on the coefficient matrix $\boldsymbol{Q}$, and does not impose non-negative constraints on the data matrix $\boldsymbol{P}$ and the basis matrix $\boldsymbol{O}$. The mathematical expression is $\boldsymbol{P}\approx\boldsymbol{O}\boldsymbol{Q}^+$, Where $\boldsymbol{Q}^+$ indicates that the matrix $\boldsymbol{Q}$ is non-negative, and the optimization goal solved by Semi-NMF is as follows:

\begin{equation}
{\min_{\boldsymbol{O},\boldsymbol{Q}}{\parallel}\boldsymbol{P}-\boldsymbol{OQ}\parallel}_F^2\ s.t.\ \boldsymbol{Q}\geq0
\label{eq2}
\end{equation}

Semi-NMF constructs a low-dimensional representation $\boldsymbol{q}$ of the original data $\boldsymbol{P}$. However, the mapping $\boldsymbol{Q}$ between this low-dimensional representation $\boldsymbol{Q}$ and the original data matrix $\boldsymbol{P}$ may contain quite complex implicit hierarchical information, which cannot be explained by a single-layer matrix factorization method. Therefore, it is necessary to mine deeper hidden information. In this case, the Deep Semi-NMF model \citep{trigeorgis2016deep} is introduced. This model adds multiple layers of factorization and decomposes the original data matrix $\boldsymbol{P}$ layer by layer into $m+1$ factors, the mathematical expression is as follows: $\boldsymbol{P}\approx\boldsymbol{O}_1\boldsymbol{O}_2\ldots\boldsymbol{O}_m\boldsymbol{Q}_m^+$, where $\boldsymbol{Q}_m^+$ is the m-level implicit representation of the data, which can be given by the following factorization.

\begin{equation}\begin{aligned}
\boldsymbol{Q}_{m-1}^+& \approx\boldsymbol{O}_m\boldsymbol{Q}_m^+ \\
\end{aligned}\label{eq3}
\end{equation}

The implicit representations $(\boldsymbol{Q}_1^+,\boldsymbol{Q}_2^+,\ldots,\boldsymbol{Q}_m^+)$ are further constrained to be non-negative. By doing so, further factorization of the low-dimensional representation can mine deeper information. The optimization objective of Deep Semi-NMF is as follows:
\begin{align}
&\min_{\boldsymbol{O,Q}} \  {\parallel} \boldsymbol{P} - \boldsymbol{O}_1 \boldsymbol{O}_2 \ldots \boldsymbol{O}_m \boldsymbol{Q}_m^+ {\parallel}_F^2 
&\text{s.t.}  \boldsymbol{Q}_m \geq 0
\label{eq4}
\end{align}

\subsection{Linear MLMF}\label{subsec2}

The optimization objective function solved by Deep Semi-NMF can be extended from single-view data to multi-omics data. The optimization objectives are as follows:

\begin{equation}
\small
\begin{aligned}
& \min_{\boldsymbol{Z}_i^{(v)}, \boldsymbol{H}_m^{(v)}} 
\sum_{v=1}^V \left(\left\| \boldsymbol{X}^{(v)} - \boldsymbol{Z}_1^{(v)} \boldsymbol{Z}_2^{(v)} \ldots \boldsymbol{Z}_m^{(v)} \boldsymbol{H}_m^{(v)} \right\|_F^2 \right. \\
& \left. + \textstyle \sum_j \left\| (\boldsymbol{H}_m^{(v)})_{.j} \right\|_1^2 \right) \\
& \text{s.t.} \boldsymbol{H}_m^{(v)} \geq 0
\end{aligned}
\label{eq5}
\end{equation}
Among them, $\boldsymbol{H}_m^{\left(v\right)}$ represents the $\textit{m}$ layer implicit of the $\textit{v}$ omics data, and the $\sum_j\left\|(\boldsymbol{H}_m^{(v)})_{.\mathrm{\textit{j}}}\right\|_1^2 $ module is used to control the sparsity of $\boldsymbol{H}_m^{\left(v\right)}$, and the specific formula is as follows:
\begin{equation}
\textstyle\sum_j\left\|\left(\boldsymbol{H}_m^{(v)}\right)_{\cdot j}\right\|_1^2=Tr\left[\left(\boldsymbol{H}_m^{(v)}\right)\left(\boldsymbol{H}_m^{(v)}\right)^T\boldsymbol{E}\right]
\label{eq6}
\end{equation}
$\boldsymbol{E}$ is a matrix with all elements equal to 1, and $Tr(\cdot)$ represents the trace operation of the matrix. Different from the common feature fusion method, the method proposed here first randomly initializes a consensus representation $\boldsymbol{H}$, and then represents the feature data of each perspective based on the consensus representation. The mathematical expression is as follows:
\begin{equation}
{\hat{\boldsymbol{H}}}_m^{\left(v\right)}=\boldsymbol{H}\boldsymbol{G}^{\left(v\right)}
\label{eq7}
\end{equation}
Among them, $\boldsymbol{G}^{(v)}$ is the index matrix that records the missing data. By minimizing the reconstruction error, the purpose of optimizing the consensus representation $\boldsymbol{H}$ and the deep feature matrix $\boldsymbol{H}_m^{\left(v\right)}$ of each omics data can be achieved. So the optimization goal of the reconstruction stage is defined as follows:
\begin{equation}
\min_{\boldsymbol{H}_m^{(v)},\boldsymbol{H}} \sum_{v=1}^{V} \left\| \boldsymbol{H}_m^{(v)} - \boldsymbol{H}\boldsymbol{G}^{(v)} \right\|_F^2
\label{eq8}
\end{equation}

To sum up, the overall optimization object of linear MLMF can be written as: 
\begin{equation}
\begin{aligned}
    & \min_{\boldsymbol{Z}_i^{(v)},\boldsymbol{H}_m^{(v)},\boldsymbol{H}} \sum_{v=1}^V \left( \left\|\boldsymbol{X}^{(v)} - \boldsymbol{Z}_1^{(v)}\boldsymbol{Z}_2^{(v)} \dots \boldsymbol{Z}_m^{(v)} \boldsymbol{H}_m^{(v)} \right\|_F^2 \right. \\
    & \left. + \lambda_1 \textstyle \sum_j \left\| \left( \boldsymbol{H}_m^{(v)} \right)_{.j} \right\|_1^2 
    + \lambda_2 \left\| \boldsymbol{H}_m^{(v)} - \boldsymbol{HG}^{(v)} \right\|_F^2 \right) \\
    & \text{s.t. } \boldsymbol{H}_m^{(v)} \geq 0
\end{aligned}
\label{eq9}
\end{equation}
Among them, $\lambda_1$ and $\lambda_2$ are penalty trade-off coefficients.

The problem is solved using the gradient descent method, which iteratively updates the variables to minimize the optimization objective function of MLMF. In each iteration, the parameter value is adjusted in the negative gradient direction according to the gradient information of the objective function relative to the parameter, and the step size is determined by the learning rate. This process continues until it converges to a local minimum or meets the stopping condition. The detailed solution process for each variable is shown in the Supplementary Note 1.

For $\boldsymbol{Z}_i^{(v)}\ (1\le i\le m)$, it is updated as follows:
\begin{equation}
\boldsymbol{Z}_i^{\left(v\right)}=\boldsymbol{X}\boldsymbol{\phi}^{-1}\boldsymbol{X}^{\left(v\right)}(\boldsymbol{H}_i^{\left(v\right)})^{-1}\ 
\label{eq10}
\end{equation}
where $\boldsymbol{\phi}=\boldsymbol{Z}_1^{\left(v\right)}\boldsymbol{Z}_2^{\left(v\right)}...\boldsymbol{Z}_{i-1}^{(v)}$ and $\boldsymbol{H}_i^{(v)}=\boldsymbol{Z}_{i+1}^{(v)}...\boldsymbol{Z}_m^{(v)}\boldsymbol{H}_m^{(v)}$.

For $\boldsymbol{H}_m^{(v)}$, it is updated as follows:
For $\boldsymbol{Z}_i^{(v)}\ (1\le i\le m)$, it is updated as follows:
\begin{equation}
\boldsymbol{A}_{ik}\gets\boldsymbol{A}_{ik}\sqrt{\frac{\boldsymbol{B}_{ik}^++(\boldsymbol{C}^-\boldsymbol{A})_{ik}}{\boldsymbol{B}_{ik}^-+(\boldsymbol{C}^+\boldsymbol{A})_{ik}}}
\label{eq11}
\end{equation}
where $\boldsymbol{A}=\left(\boldsymbol{H}_m^{\left(v\right)}\right)$, and $\boldsymbol{I}$ is the unit matrix. $\boldsymbol{B}=\boldsymbol{\Psi}^T\boldsymbol{X}^{\left(v\right)}+\lambda_2\boldsymbol{H}\boldsymbol{G}^{\left(v\right)}$, $\boldsymbol{C}=\boldsymbol{\Psi}^T\boldsymbol{\Psi}+\lambda_1\boldsymbol{E}+\lambda_2\boldsymbol{I}$.

For $\boldsymbol{H}_i^{\left(v\right)}\ (i<m)$, it is updated as follows:
\begin{equation}
\boldsymbol{H}_{ik}^{\left(v\right)} \gets \boldsymbol{H}_{ik}^{\left(v\right)} 
\sqrt{\frac{(\boldsymbol{\Psi}^T \boldsymbol{X}^{\left(v\right)})_{ik}^+ 
+ \left(\left(\boldsymbol{\Psi}^T \boldsymbol{\Psi}^-\right)\boldsymbol{H}\right)_{ik}} 
{(\boldsymbol{\Psi}^T \boldsymbol{X}^{\left(v\right)})_{ik}^- 
+ \left(\left(\boldsymbol{\Psi}^T \boldsymbol{\Psi}^-\right)\boldsymbol{H}\right)_{ik}}}
\label{eq12}
\end{equation}
where $\boldsymbol{\Psi}=\boldsymbol{Z}_1^{\left(v\right)}\boldsymbol{Z}_2^{\left(v\right)}\ldots\boldsymbol{Z}_i^{\left(v\right)}$. 

For $\boldsymbol{H}$, it is updated as follows:
\begin{equation}
\boldsymbol{H} = \sum_{v=1}^{V} \; \boldsymbol{H}_m^{(v)} \boldsymbol{G}^{(v)T} \left( \sum_{v=1}^{V} \; \boldsymbol{G}^{(v)} \boldsymbol{G}^{(v)T} \right)^{-1}
\label{eq13}
\end{equation}

Summarizing the above steps, the optimization process of the Linear MLMF is shown in Algorithm \ref{algo1}.

\begin{algorithm}[!t]
\caption{Algorithm of Linear MLMF}\label{algo1}
\begin{algorithmic}[1]
\Statex \hspace{-0.5cm}\textbf{Input:} multi-omics data $\boldsymbol{X}$, trade-off coefficients $\lambda_1$, $\lambda_2$ 
\Statex \hspace{-0.5cm}\textbf{Output:} consensus representation $\boldsymbol{H}$ 
\State Construct an indicator matrix $\boldsymbol{G}^{(v)}$ via Eq. (\ref{eq1})
\State Initialize each matrix
\State Update $\boldsymbol{Z}_i^{(v)}$ according to Eq. (\ref{eq10})
\State Update $\boldsymbol{H}_m^{(v)}$ according to Eq. (\ref{eq11})
\State Update $\boldsymbol{H}_i^{(v)} (i < m)$ according to Eq. (\ref{eq12})
\State Update $\boldsymbol{H}$ according to Eq. (\ref{eq13})
\State Repeat steps 3-6 until convergence
\State Return $\boldsymbol{H}$
\end{algorithmic}
\end{algorithm}

\subsection{Nonlinear MLMF}\label{subsec2}

By linearly decomposing the initial data distribution, it may not be possible to effectively describe the nonlinear relationship between the potential attributes of the model. Introducing nonlinear functions between layers can extract features for each potential attribute of the model, and the nonlinear functions are nonlinearly separable in the initial input space. After constructing the optimization target, the gradient descent method is used to solve it. 

First, construct the loss function. Compared with linear factorization, nonlinear factorization uses nonlinear mapping in all factorizations except the first layer. Nonlinear factorization decomposes the given data matrix $\boldsymbol{X}$ into $m+1$ factors in a nonlinear way, as $\boldsymbol{X}\approx\boldsymbol{Z}_1f(\boldsymbol{Z}_2f(...f(\boldsymbol{Z}_m\boldsymbol{H}_m^+)))$. $\boldsymbol{H}_m^+$ is the m-level implicit representation of the data, which can be given by the following factorization: 

\begin{equation}\begin{aligned}
\boldsymbol{H}_{m-1}^+& \approx f(\boldsymbol{Z}_m\boldsymbol{H}_m^+) \\
\end{aligned}\label{eq14}\end{equation}

The optimization goal of the deep matrix nonlinear factorization model is as follows:

\begin{equation}
\small
\begin{aligned}
L = & \min_{\boldsymbol{Z}_i^{(v)},\boldsymbol{H}_m^{(v)},\boldsymbol{H}} 
\sum_{v=1}^{V} \left\| \boldsymbol{X}^{(v)} - \boldsymbol{Z}_{1}^{(v)} 
f \left( \boldsymbol{Z}_{2}^{(v)} f \left( \ldots f \left( 
\boldsymbol{Z}_{m}^{(v)} \boldsymbol{H}_{m}^{(v)} \right) \right) \right) 
\right\|_{F}^{2} \\
& \quad + \lambda_1 \textstyle \sum_{j} \left\| \left( \boldsymbol{H}_m^{(v)} \right)_{.j} \right\|_1^{2} 
+ \lambda_2 \left\| \boldsymbol{H}_m^{(v)}-\boldsymbol{HG}^{(v)} \right\|_{F}^{2} \\
& \text{s.t.} \boldsymbol{H}_m^{(v)} \geq 0
\end{aligned}
\label{eq15}
\end{equation}

The problem is solved using the gradient descent method, which iteratively updates the variables to minimize the optimization objective function of MLMF. In each iteration, the parameter value is adjusted in the negative gradient direction according to the gradient information of the objective function relative to the parameter, and the step size is determined by the learning rate. This process continues until it converges to a local minimum or meets the stopping condition. The detailed solution process for each variable is shown in the Supplementary Note 2.

For $\boldsymbol{H}_i^{(v)}\ (1\le i\le m)$, it is updated as follows:

\begin{equation}
\boldsymbol{H}_i^{\left(v\right)}=\boldsymbol{H}_i^{\left(v\right)}-\alpha\frac{\partial L}{\partial\boldsymbol{H}_i^{\left(v\right)}}
\label{eq16}\end{equation}
where $\boldsymbol{H}_i^{\left(v\right)}=\boldsymbol{H}_{i+1}^{\left(v\right)}\boldsymbol{Z}_{i+1}^{\left(v\right)}$

For $\boldsymbol{Z}_i^{(v)}\ (1\le i\le m)$, it is updated as follows:

\begin{equation}
\boldsymbol{Z}_i^{\left(v\right)}=\boldsymbol{Z}_i^{\left(v\right)}-\alpha\frac{\partial L}{\partial\boldsymbol{Z}_i^{\left(v\right)}}
\label{eq17}\end{equation}

For $\boldsymbol{H}$, it is updated as follows:

\begin{equation}
\boldsymbol{H}=\boldsymbol{H}-\alpha\frac{\partial L}{\partial\boldsymbol{H}}
\label{eq18}\end{equation}

To summarize the above steps, each variable is regarded as the only variable of the objective function, and its partial derivative is taken as the gradient. The variable is updated using the gradient descent method. The optimal solution is obtained by alternately updating the variables. The optimization process of the deep matrix nonlinear factorization algorithm is shown in Algorithm \ref{algo2}.

\begin{algorithm}[!t]
\caption{Algorithm of Nonlinear MLMF}\label{algo2}
\begin{algorithmic}[1]
\Statex \hspace{-0.5cm}\textbf{Input:} multi-omics data $\boldsymbol{X}$, trade-off coefficient $\lambda_1$, $\lambda_2$, step length $\alpha$.
\Statex \hspace{-0.5cm}\textbf{Output:} consensus representation $\boldsymbol{H}$ 
\State Construct an indicator matrix $\boldsymbol{G}^{(v)}$ via Eq. (\ref{eq1})
\State Initialize each matrix
\State Update $\boldsymbol {H}_i^{(v)} (i < m)$ according to Eq. (\ref{eq16})
\State Update $\boldsymbol {Z}_i^{(v)} (i < m)$ according to Eq. (\ref{eq17})
\State Update $\boldsymbol {H}$ according to Eq. (\ref{eq18})
\State Repeat steps 3-8 until convergence
\State Return $\boldsymbol {H}$
\end{algorithmic}
\end{algorithm}

\subsection{Spectral clustering}\label{subsec2}

The consensus representation $\boldsymbol{H}$ is clustered using the spectral clustering method (\citep{von2007tutorial}). First, a similarity matrix is constructed. This paper uses the k-nearest neighbor method to build the similarity matrix, expressed as follows:

\begin{equation}
\boldsymbol{W}_{ij} =
\begin{cases}
    0, & \boldsymbol{h}_i \notin nei(\boldsymbol{h}_j) \text{ and } \boldsymbol{h}_j \notin nei(\boldsymbol{h}_i) \\
    \exp\left( -\frac{\|\boldsymbol{h}_i - \boldsymbol{h}_j\|^2}{2\sigma^2} \right), & \boldsymbol{h}_i \in nei(\boldsymbol{h}_j) \text{ or } \boldsymbol{h}_j \in nei(\boldsymbol{h}_i)
\end{cases}
\label{19}
\end{equation}
where $\sigma$ is a tuning parameter to scale the similarity measure. The standardized Laplace matrix can be obtained as follows:

\begin{equation}
\boldsymbol{L}=\boldsymbol{D}^{-\frac{1}{2}}\boldsymbol{W}\boldsymbol{D}^{-\frac{1}{2}}
\label{20}
\end{equation}
among them, $\boldsymbol{D}$ is the diagonal matrix of $\boldsymbol{W}$, calculated as $\boldsymbol{D}_{ii}=\sum_{ij}\boldsymbol{W}_{ij}$.

The third step is to to optimize the following objective function based on the Laplacian matrix $\boldsymbol{L}$:

\begin{equation}
\begin{aligned}&\min_{{\boldsymbol{B}}}Tr(\boldsymbol{B}^{T}\boldsymbol{L}\boldsymbol{B})\\&s.t.\boldsymbol{B}^{T}\boldsymbol{B}=\boldsymbol{I}\end{aligned}
\label{21}
\end{equation}
where $\boldsymbol{B}$ is the indicator matrix, defined as $\boldsymbol{B}=\boldsymbol{Y}(\boldsymbol{Y}^T\boldsymbol{Y})^{-\frac{1}{2}}$. Among them, $\boldsymbol{Y}=[\boldsymbol{y}_1^T,\boldsymbol{y}_2^T,...,\boldsymbol{y}_n^T]^T$, $\boldsymbol{y}_i=[\boldsymbol{y}_{i1},\boldsymbol{y}_{i2},...,\boldsymbol{y}_{ik}]$ is the clustering result, $\boldsymbol{y}_{ik}=1$ means that the i-th sample belongs to the k-th class. $\boldsymbol{I}$ is the identity matrix, and the constraint $\boldsymbol{B}^T\boldsymbol{B}=\boldsymbol{I}$ is to control each sample to belong to only one category. So the optimization problem is transformed into finding the eigenvectors corresponding to the first $k$ smallest eigenvalues of the graph Laplacian matrix $L$. Then, the matrix $\boldsymbol{B}=[\boldsymbol{b}_1,\boldsymbol{b}_2,...,\boldsymbol{b}_k]$ is treated as a new data set with k-dimensional features and $n$ samples for K-Means clustering, and the category to which each sample belongs can be obtained.

\section{Results}\label{sec4}
\subsection{Full muti-omics datasets}\label{subsec2}

\begin{table*}[t]
\caption{The comparison of clustering results from different algorithms on ten simulated full TCGA dataset\label{tab1}}
\tabcolsep=0pt
\begin{tabular*}{\textwidth}{@{\extracolsep{\fill}}lccccccccccc@{\extracolsep{\fill}}}
\toprule%
Alg./Cancer & AML & BIC & COAD & GBM & KIBC & LIHC & LUSC & OV & SKCM & SARC & Mean\\
\midrule
K-means & \textbf{1}/\textbf{2.4} & \textbf{2}/\textbf{3.5} & \textbf{1}/0.4 & \textbf{2}/\textbf{2.6} & \textbf{1}/0.8 & \textbf{2}/0.2 & 0/\textbf{1.5} & \textbf{2}/0.3 & \textbf{2}/0.9 & \textbf{2}/\textbf{1.3} & \textbf{1.5}/\textbf{1.5}\\
Spectral & \textbf{1}/\textbf{2.1} & \textbf{1}/\textbf{5.0} & \textbf{1}/0.7 & \textbf{2}/\textbf{2.5} & \textbf{2}/\textbf{1.8} & \textbf{2}/0.4 & 0/\textbf{2.1} & \textbf{2}/0.8 & 0/0.6 & \textbf{2}/\textbf{1.3} & \textbf{1.3}/\textbf{1.7}\\
LRAcluster & \textbf{1}/\textbf{1.8} & \textbf{2}/\textbf{4.0} &\textbf{ 1}/0.1 & \textbf{2}/1.1 & \textbf{2}/1.0 & \textbf{2}/\textbf{2.4} & \textbf{1}/1.0 & \textbf{2}/0.2 & \textbf{3}/\textbf{2.9} & \textbf{2}/\textbf{2.5} & \textbf{1.8}/\textbf{1.7}\\
CC & \textbf{1}/\textbf{3.8} & \textbf{1}/\textbf{2.8} & \textbf{1}/0.5 & \textbf{2}/\textbf{2.1} & \textbf{3}/\textbf{1.3} & \textbf{2}/0.5 & \textbf{1}/1.1 & \textbf{1}/0.2 & \textbf{3}/\textbf{2.5} & \textbf{2}/1.0 & \textbf{1.5}/\textbf{1.4}\\
PINS & \textbf{1}/\textbf{1.6} & \textbf{1}/\textbf{2.8} & 0/0.5 & \textbf{1}/\textbf{4.4} & \textbf{2}/1.0 & \textbf{2}/0.8 & 0/\textbf{1.9} & \textbf{1}/0.1 & \textbf{1}/\textbf{1.0} & \textbf{2}/0.8 & \textbf{1.1}/\textbf{1.5}\\
MCCA & \textbf{1}/1.2 & \textbf{1}/\textbf{8.0} & 0/0.2 & \textbf{1}/\textbf{2.9} & \textbf{2}/\textbf{1.8} & \textbf{2}/1.1 & \textbf{2}/\textbf{2.3} & 0/0.6 & \textbf{2}/\textbf{4.7} & \textbf{2}/\textbf{1.5} & \textbf{1.3}/\textbf{2.4}\\
iClusterBayes & \textbf{1}/\textbf{1.5} & 0/\textbf{1.3} & \textbf{2}/0.1 & \textbf{1}/\textbf{3.1} & \textbf{4}/\textbf{7.3} & \textbf{2}/\textbf{2.2} & 0/\textbf{1.5} & \textbf{2}/0.9 & \textbf{2}/0.6 & \textbf{2}/\textbf{3.7} & \textbf{1.6}/\textbf{2.2}\\
SNF & \textbf{1}/\textbf{3.0} & \textbf{2}/\textbf{6.0} & \textbf{1}/0.2 & \textbf{2}/\textbf{2.6} & \textbf{3}/\textbf{1.7} & \textbf{2}/0.3 & \textbf{1}/1.2 & \textbf{2}/0.2 & \textbf{1}/1.1 & \textbf{2}/\textbf{1.9} & \textbf{1.5}/\textbf{1.9}\\
SNFCC & \textbf{1}/\textbf{3.8} & \textbf{3}/\textbf{7.2} & \textbf{2}/0.6 & \textbf{2}/\textbf{2.3} & \textbf{2}/1.1 & \textbf{1}/1.2 & \textbf{1}/1.0 & \textbf{1}/0.2 & \textbf{2}/0.6 & \textbf{2}/1.1 & \textbf{1.8}/\textbf{2.1}\\
NEMO & \textbf{1}/\textbf{1.8} & \textbf{2}/\textbf{4.2} & 0/0.1 & \textbf{1}/\textbf{3.8} & \textbf{4}/\textbf{2.2} & \textbf{4}/\textbf{4.2} & 0/\textbf{1.8} & \textbf{1}/0.4 & \textbf{3}/\textbf{4.0} & \textbf{2}/\textbf{1.9} & \textbf{1.7}/\textbf{2.4}\\
IntNMF & \textbf{1}/\textbf{1.9} & \textbf{1}/\textbf{4.3} & \textbf{1}/0.2 & \textbf{1}/\textbf{3.5} & \textbf{3}/0.2 & \textbf{2}/\textbf{2.0} & 0/0.9 & 0/0.7 & \textbf{2}/\textbf{4.1} & \textbf{2}/\textbf{1.8} & \textbf{1.8}/\textbf{2.1}\\
MLMF\_Linear & \textbf{1}/\textbf{3.4} & \textbf{3}/\textbf{5.9} & \textbf{1}/0.4 & \textbf{2}/\textbf{4.1} & \textbf{2}/\textbf{1.4} & \textbf{2}/\textbf{3.2} & \textbf{1}/\textbf{1.8} & \textbf{2}/\textbf{1.9} & \textbf{3}/\textbf{2.9} & \textbf{2}/1.0 & \textbf{1.9}/\textbf{2.6}\\
MLMF\_Noninear & \textbf{1}/\textbf{3.1} & \textbf{4}/\textbf{5.5} & \textbf{2}/0.3 & \textbf{1}/\textbf{4.5} & \textbf{3}/\textbf{1.5} & \textbf{3}/\textbf{3.1} & \textbf{1}/\textbf{1.6} & \textbf{1}/\textbf{2.7} & \textbf{3}/\textbf{4.3} & \textbf{2}/0.8 & \textbf{2.1}/\textbf{2.7}\\
\botrule
\end{tabular*}
\begin{tablenotes}%
\item Note: in each cell A/B, A is significant clinical parameters detected. B is -log10 P-value for survival. 0.05 is the threshold for significance and the bold indicates the significant results. Mean is algorithm average value.
\end{tablenotes}
\end{table*}

\begin{figure}[!t]
\centering
\includegraphics[width=0.35\textwidth]{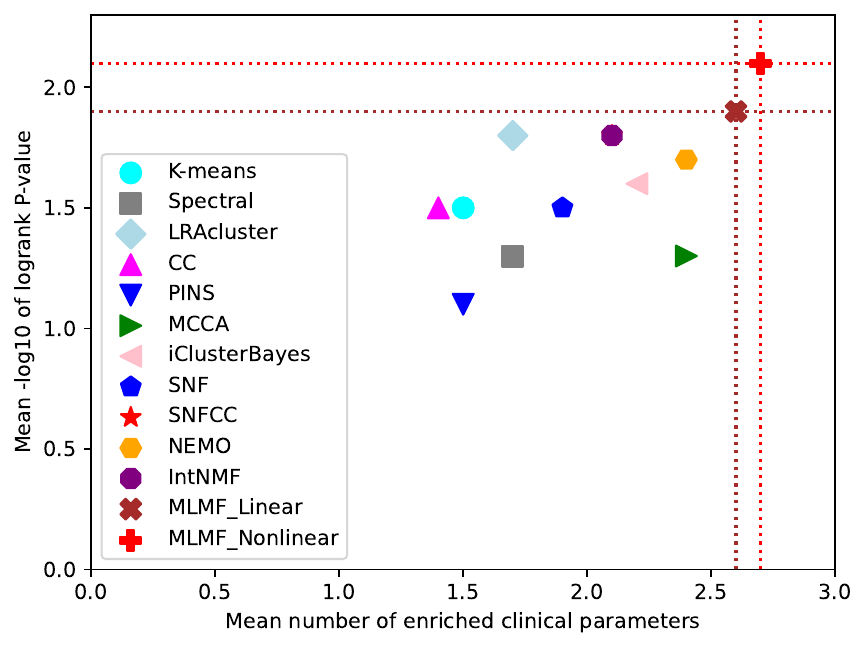} 
\caption{Mean performance of the different algorithms on 10 cancer datasets. Y-axis represents average -log10 logrank test’s P-values and X-axis represents average number of enriched clinical parameters in the clusters. The red dotted lines highlight the results of MLMF\_Nonlinear and the brown dotted lines highlight the results of MLMF\_Linear.}
\label{fig2}
\end{figure}



Several computational experiments evaluate the effectiveness of cancer subtypes with multi-omics data. This paper conducts experiments on 10 cancer data sets of AML, BIC, COAD, GBM, KIRC, LIHC, LUSC, OV, SKCM and SARC of TCGA \citep{cancer2008comprehensive}. Each data set includes mRNA expression, DNA methylation and miRNA expression data. The feature data after dimensionality reduction is standardized using z-score. 

This article compares MLMF with ten algorithms are selected as comparisons methods on complete multi-omics data sets, including K-means and spectral clustering algorithms, as well as eight integration methods such as LRAcluster \citep{wu2015fast}, PINS (\citep{nguyen2017novel}, MCCA \citep{witten2009extensions}, iClusterBayes \citep{mo2018fully}, SNF \citep{wang2014similarity}, SNFCC \citep{xu2017cancersubtypes}, NEMO \citep{rappoport2019nemo}, and IntNMF \citep{hosmer2008applied}. The evaluation indicators used for the identified subtype results are the enrichment number of clinical parameters and the significance of survival analysis. The number of subtypes for each cancer type was determined by feature factorization. For simplicity, both the penalty coefficients $\lambda_1$ and $\lambda_2$ are set to 1, and the step size is adjusted in an adaptive manner. Survival analysis using the Cox proportional hazards model and p-value showed statistically significant differences in the survival spectra of different cancer subtypes \citep{yang2022subtype}. To perform enrichment analysis of clinical signatures, we selected a unified set of patient clinical information for all cancers, such as sex and age at initial diagnosis, as well as quantifying tumor progression (pathology T), lymph node cancer (pathology N), metastasis (pathology M) and overall progression (pathological stage) as four discrete clinicopathological parameters \citep{ding2008convex}. Following the recommendations of Rappoport and Shamir (2019), the number of clusters in the comparison method was set to the same value as reported in the original paper.

 Table \ref{tab1} and  Figure \ref{fig2} show the cancer subtype prediction performance of different algorithms on 10 complete TCGA data sets. As can be seen from the results, the clusters discovered by MLMF\_Linear had significant survival differences in 9 of the 10 cancer datasets, and the clusters discovered by MLMF\_nonLinear had significant survival differences in 8 of the 10 cancer datasets. The average logrank p-value of MLMF\_Nonlinear reaches 2.7, and the average logrank p-value of MLMF\_Linear reaches 2.6. MCCA and NEMO ranked third with 1.8. None of the methods found significant differences in survival rates for the COAD dataset. MLMF\_Linear and MLMF\_Nonlinear found at least one enriched clinical parameter in all datasets. The average number of enriched clinical parameters for MLMF\_Nonlinear was 2.1, and the average number of enriched clinical parameters for MLMF\_Linear was 1.9. These results show that linear factorization and nonlinear factorization of MLMF can identify patient subtypes with significant consistency and clinical relevance, and the overall effect of nonlinear factorization is slightly higher than that of linear factorization.
 
In order to verify the subtypes obtained by MLMF\_Linear and the existing subtypes, and to show the differential expression between different subtypes, this paper designed the following experiments. First, the subtype results of PAM50 on the BIC dataset were selected for comparison. Secondly, since there were 48 mRNA expression features associated with the 50 genes of PAM50, we deleted the 48 features in the original mRNA data of the BIC dataset to eliminate the direct effects of known oncogenes in multi-omics data, and then input the processed mRNA data into MLMF\_Linear together with other omics data. Finally, a heat map was drawn using the expression of the 48 mRNAs to show the correlation between oncogenes and subtypes obtained from MLMF\_Linear, as well as the overlap of subtypes obtained by MLMF\_Linear and PAM50. As shown in Supplementary Fig S1, different subtypes have different mRNA expression patterns, and there is a large overlap between MLMF\_Linear and PAM50, such as the LumA subtype of PAM and subtype 1 of MLMF\_Linear, and the Basal subtype of PAM and subtype 3 of MLMF\_Linear.

In order to verify the training effect of the MLMF algorithm, this paper records the changes in the loss function values of MLMF\_Linear and MLMF\_Nonlinear under 20 epochs, as shown in Supplementary Fig S2. It can be seen from the figure that the loss of MLMF\_Linear and MLMF\_Nonlinear both show a downward and convergent trend. MLMF\_Linear has a great improvement in the early stage of training, and the loss drops rapidly. The convergence process of MLMF\_Nonlinear is more stable, showing a gradual downward trend.

\subsection{Partial multi-omics datasets}\label{subsec2}

To evaluate the performance of the method on some multi-omics datasets, this paper still selected the ten TCGA datasets analyzed above and simulated some patient loss omics measurements. Specifically, this paper maintains the complete expression of DNA methylation and miRNA, and randomly extracts samples from a part of patients to remove their mRNA expression, with missing rates of 0.1, 0.3, 0.5, and 0.7. Enrichment analysis and survival analysis are still used to evaluate the performance of the method. Supplementary Table 1 shows the comparison results of different algorithms on ten simulated missing TCGA datasets.

From Supplementary Table 1 and Supplementary Fig S3, MLMF\_Linear and MLMF\_Nonlinear performed better than NEMO and MCCA in survival and enrichment analysis at all missing rates. Under the same missing rate, the average performance of the nonlinear decomposition algorithm is better than that of the linear decomposition. These results indicate that MLMF can be well applied to situations where part of the omics is missing. In general, cancer subtyping by MLMF resulted in statistically significant survival spectrum differences and significant clinical enrichment. In addition, MLMF can effectively solve the challenge of missing parts of the omics.

In order to evaluate the efficiency of the MLMF algorithm,we compared the average running time of the MLMF\_Linear algorithm and the MLMF\_Nonlinear algorithm on the BIC dataset with ten algorithms, namely K-means, spectral clustering algorithms, LRAcluster, PINS, MCCA, iClusterBayes, SNF, SNFCC, NEMO. As can be seen from Supplementary Fig S4, the fastest algorithm is spectral clustering and the slowest algorithm is iClusterBayes. In general, the running time of the MLMF algorithm saves more time than the training model of the deep neural network, and the results are better than those of the ordinary clustering algorithm.

\section{Conclusion}

Predicting cancer subtypes using multi-omics data enables researchers and clinicians to adopt a more comprehensive and precise approach to patient treatment. Data from various omics offer distinct insights into biological processes, and by integrating these multi-omics datasets, researchers can uncover unique patterns and molecular features associated with different cancer subtypes. In this paper, we introduce MLMF, a multi-layer matrix decomposition method designed for cancer subtyping through the clustering of multi-omics data. For the first time, MLMF unifies the processing pipelines for complete and missing multi-omics data within a common framework. It performs multi-layer linear or nonlinear decomposition on the multi-omics feature matrix, breaking down the original data representation into respective latent feature representations. These representations are then fused to create a consensus representation. The identification of cancer subtypes is achieved through spectral clustering of this consensus representation. Experimental results from 10 TCGA multi-omics datasets demonstrate that MLMF outperforms other related methods. While our study focused on two to three histological levels, MLMF provides a versatile framework that can be easily adapted to scenarios involving additional omics data. We believe that MLMF holds significant promise for advancing precision oncology and enhancing patient outcomes.

\bibliographystyle{apalike}
\bibliography{reference}


\end{document}